\documentclass[sigconf]{acmart}

\usepackage{microtype}
\usepackage{booktabs}
\usepackage{paralist}
\usepackage{tabularx}
\usepackage{subcaption}
\usepackage{xspace}
\usepackage{tabu}

\newcommand{\xlmroberta}{\textsc{XLM-RoBERTa}\xspace}
\newcommand{\mdeberta}{\textsc{mDeBERTa}\xspace}
\newcommand{\ernie}{\textsc{Ernie}\xspace}
\newcommand{\minilm}{\textsc{MiniLMv2}\xspace}
\newcommand{\xlmv}{\textsc{XLM-V}\xspace}
\newcommand{\mdebertatasksource}{\textsc{mDeBERTa-TS}\xspace}
\newcommand{\ie}{\textit{i.e.}\xspace}
\newcommand{\eg}{\textit{e.g.}\xspace}
\newcommand{\F}{F$_1$\xspace}
\newcommand{\heatmapscale}{0.525}
\newcommand{\citewithname}[1]{\citeauthor{#1}~\cite{#1}}

\copyrightyear{2024}
\acmYear{2024}
\setcopyright{acmlicensed}
\acmConference[WWW '24 Companion]{Companion Proceedings of the ACM Web Conference 2024}{May 13--17, 2024}{Singapore, Singapore}
\acmBooktitle{Companion Proceedings of the ACM Web Conference 2024 (WWW '24 Companion), May 13--17, 2024, Singapore, Singapore}
\acmDOI{10.1145/3589335.3651902}
\acmISBN{979-8-4007-0172-6/24/05}

\begin{document}

\title{English Prompts are Better for NLI-based\\ Zero-Shot Emotion Classification than Target-Language Prompts}

\author{Patrick Barei\ss}
\affiliation{%
  \institution{University of Stuttgart}
  \country{Germany}}
\email{patrick.bareiss@ims.uni-stuttgart.de}

\author{Roman Klinger}
\affiliation{%
  \institution{University of Bamberg}
  \country{Germany}}
\email{roman.klinger@uni-bamberg.de}

\author{Jeremy Barnes}
\affiliation{%
  \institution{University of the Basque Country}
  \country{Spain}}
\email{jeremy.barnes@ehu.eus}

\begin{abstract}
  Emotion classification in text is a challenging task due to the
  processes involved when interpreting a textual description of a
  potential emotion stimulus. In addition, the set of emotion
  categories is highly domain-specific. For instance, literature
  analysis might require the use of aesthetic emotions (e.g., finding
  something beautiful), and social media analysis could benefit from
  fine-grained sets (e.g., separating anger from annoyance) than only
  those that represent basic categories as they have been proposed by
  Paul Ekman (anger, disgust, fear, joy, surprise, sadness). This
  renders the task an interesting field for zero-shot classifications,
  in which the label set is not known at model development
  time. Unfortunately, most resources for emotion analysis are
  English, and therefore, most studies on emotion analysis have been
  performed in English, including those that involve prompting
  language models for text labels. This leaves us with a research gap
  that we address in this paper: In which language should we prompt
  for emotion labels on non-English texts? This is particularly of
  interest when we have access to a multilingual large language model,
  because we could request labels with English prompts even for
  non-English data. Our experiments with natural language
  inference-based language models show that it is consistently better
  to use English prompts even if the data is in a different language.
\end{abstract}



\maketitle

\section{Introduction}
Pretraining large language models (LLMs) on large amounts of text and
subsequently fine-tuning them for a specific task constitutes a
\textit{de facto} state of the art to address several natural language
processing (NLP) tasks, \eg, sentiment analysis
\cite{yang2019xlnet,sun2019}, question answering \cite{raffel2020},
or natural language inference
\cite{raffel2020,zhuang-etal-2021-robustly}.
This includes emotion classification, a popular and important task
with many datasets from various domains
\cite[i.a.]{bostan-klinger-2018-analysis,mohammad-bravo-marquez-2017-emotion,li-etal-2017-dailydialog,scherer-wallbott-1994-emotion}.

Most work on emotion analysis has been performed in English
\citep[see][]{bostan-klinger-2018-analysis}, although there has been
some work in other languages
\citep[i.a.]{schmidt-etal-2021-emotion,Troiano2019,Cheng2017}. However,
the difficulty and high cost of annotating a large emotion
classification dataset means that most languages do not have any
resources available. In such a situation, zero-shot cross-lingual
methods are of interest.

Driven by the increasing abilities of LLMs to generalize across tasks,
recent research has shifted away from fine-tuning models for each new
task, instead focusing on zero and few-shot learning
\cite{song2023comprehensive, zhao2023survey}, and oftentimes
reformulating the original tasks as natural language inference (NLI)
\cite{bowman-etal-2015-large}. This approach enables the use of a
language model that has been fine-tuned on an NLI dataset to perform a
new task without further tuning the model
\citep{schick-schutze-2021-shot,schick-schutze-2021-just}. This
reformulation can be done programatically, creating and filling prompt
templates that correspond to NLI premises and hypotheses. Such
zero-shot classification can achieve good results
\citep{yin-etal-2019-benchmarking}, including emotion classification
\citep{plaza-del-arco-etal-2022-natural}.

\begin{figure}[b]
  \centering
  \includegraphics[width=0.9\linewidth]{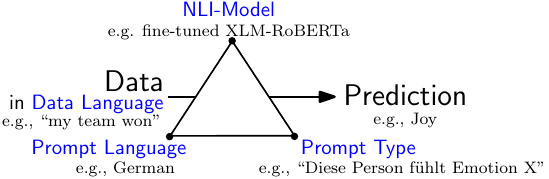}
  \caption{We study the interaction of \textit{data} and
    \textit{prompt language}, while considering the underlying
    \textit{NLI-model} and the role of the \textit{prompt type}.}.
    \label{fig:variable_explanation}
\end{figure}

Such NLI-based approach to emotion classification checks if a specific
sentence entails information of the classification instance using the
prompt. For instance, given a sentence ``I won in the lottery'', an
NLI model shall return a high entailment probability for the prompt
``This sentence expresses \underline{joy}'' but a low probability for
``This sentence expresses~\underline{anger}''. We assume the standard
setup for zero-shot classification using NLI, in which the model is
not further fine-tuned for emotion classification.

In the established supervised learning regime,
obtaining models for a low-resource target-language that is different from the language of the available training data, \ie cross-lingual model induction, has been approached commonly by either 1)
transforming the data in some way to create target-language data --
oftentimes using translation or label projection, or 2) using model
transformations to create a language-agnostic model.

However as many NLI models are inherently multilingual, they can perform a task in a low-resource target-language without additional training when used in a zero-shot manner and thus inducing training data in the target-language or making the model language-agnostic is unnecessary. Instead, the object of focus for cross-lingual transfer shifts to identifying the prompt most optimally suited for classifying data in the target-language. As the
prompt does not need to be in the same language as the data are
in, one approach is to use an existing known
well-performing prompt in a high-resource and well-studied language such as
English directly. On the one hand this makes sense as English is commonly the
most prevalent language in the training data of multilingual models
and is thus likely a prompt written in it will perform well. On the
other hand it also appears sensible to match the prompt language with
the data language as common multilingual datasets used for training
NLI models (such as XNLI \cite{conneau-etal-2018-xnli}) only contain
matched examples, \eg, German prompts with German data and thus a
mismatch would be out-of-distribution for the training data and
potentially results in worse performance. To address this, a well
performing prompt in English could be translated to the data
language. But then it still remains unclear if the kind of phrasing
used to specify the prompt in the original language will be equally as
useful in the target language. Especially for emotion classification
different words can carry slightly different connotations in different
languages.  Right now answering these questions of optimal
cross-lingual prompt transfer is relatively unexplored for most tasks
\cite{zhao-schutze-2021-discrete}, with no related research available
concerning cross-lingual emotion classification.

Therefore this paper aims at answering the following question: \textit{How do we
  best transfer prompts for zero-shot emotion classification from a
  high-resource language to a low-resource language?}  We study the
relation between the \textit{data language} and the \textit{prompt
  language}, while also analyzing the impact of changes to the
\textit{prompt type} (the phrasing of the prompt) and the underlying multilingual NLI
\textit{model}. Figure~\ref{fig:variable_explanation} shows a visual
representation of this setup.
Concretely, we focus on the following research questions:
\begin{compactitem}
\item \textbf{RQ1.} Should we translate the \textit{prompt language} to match the \textit{data language} or leave it in English? \textit{(English is better)}
\item \textbf{RQ2.} Is the performance of different \textit{prompt types} stable across different \textit{data languages}? \textit{(yes)}
\item \textbf{RQ3.} How consistent are the results across different
  NLI \textit{models}? \textit{(they are consistent)}
\end{compactitem}

Our evaluation is based on 3 corpora spanning 18 languages with 7 prompt
types \cite{plaza-del-arco-etal-2022-natural} and 6 multilingual NLI
\cite{conneau-etal-2020-unsupervised,laurer2022less,sileo2023tasksource}
models.

\section{Related Work}
\subsection{Multilingual Emotion Classification}
While much early work on emotion classification in NLP focused on
English
\citep{alm-etal-2005-emotions,Mihalcea2012,mohammad-2012-emotional},
approaches and datasets to classify emotions in multiple languages,
including low-resource ones, have expanded more recently.

\citewithname{bianchi-etal-2022-xlm} collect social media emotion data
across 19 languages and use it to train an inherently multilingual
model. \citewithname{becker2017multilingual} investigates
this supervised setting with multiple experiments. Multiple labelled
multilingual emotion classification corpora exist for use in this
setting, such as Universal Joy \cite[tagged Facebook
comments,][]{lamprinidis-etal-2021-universal}, de/enISEAR
\cite[crowd-sourced self-reported event descriptions,][]{Troiano2019}
or EmoEvent \cite[tweets,][]{plaza-del-arco-etal-2020-emoevent}.
\citewithname{gupta-2021-multilingual} explores the use of multilingual
models in conjunction with unsupervised, adversarial training, \ie,
unlabelled data instead.

\citewithname{de-bruyne-2023-paradox} has pointed out problems with
such approaches, e.g., that the concept of an emotion is to some
extend dependent on the language and associated culture itself making
multilingual approaches inherently more difficult to
apply. \citewithname{de-bruyne-etal-2022-language} find evidence for
this, suggesting that typologically dissimilar languages in particular
utilize language-specific representations for classification in a
single multilingual
model. \citewithname{havaldar-etal-2023-multilingual} also investigate
this and suggest to work towards better monolingual models as well as
culturally balanced corpora for training.

\subsection{Prompt-based Learning for Emotion Classification}
Prompt-based learning for emotion classification is an attractive
alternative to more data-intensive approaches
\cite{bianchi-etal-2022-xlm}. \citewithname{plaza-del-arco-etal-2022-natural}
explore and evaluate a set of prompts extensively across
multiple corpora for this reason. Prompt-based approaches can also be
used in more complicated settings: \citewithname{yi2022contextual} propose
a prompt-based approach for emotion classification in conversation, a
task often difficult for more traditional approaches. They achieve
this by first using a language model to extract commonsense features
and use those to create a soft prompt then used for actual
classification. Another area where prompt-based learning has seen
success is in multimodel emotion classification, i.e, where the input
consists not only of text but also audio or
video. \citewithname{zhao2022memobert} use a pretrained language model in
conjunction with a prompt and combine the resulting embeddings with
data from other modalities. \citewithname{jeong2023multimodal} employ
something similar but focus only on the combination of text and
audio.  However, previous work does not evaluate these techniques
in a multilingual setting.

\subsection{Multilingual and Cross-lingual Prompting}
There is only limited work on multilingual prompting, which has,
however, shown already some promising results. As an example,
\citewithname{zhao-schutze-2021-discrete} explore few-shot
cross-lingual NLI and fine-tune multilingual LLMs with both English
and translated prompts, finding that prompting outperformed standard
supervised training in few-shot and multilingual
scenarios. \citewithname{fu-etal-2022-polyglot} experiment with
multi-task multilingual prompting on a number of tasks (summarization,
NER, QA, topic classification, sentiment analysis and NLI). They find
that training on larger amounts of available English datasets leads to
benefits for both in-language training, as well as for a cross-lingual
zero-shot scenario. They also report that training the models
uniformly on English prompts performs
better. \citewithname{huang-etal-2022-zero} find that initializing
soft prompts with embeddings taken from multilingual LLMs performs
better than translation or soft prompting with random
initialization. \citewithname{kim-komachi-2023-enhancing} concentrate
on discovering target-language examples that zero-shot prompting
cannot predict. \citewithname{nie-etal-2023-cross} instead propose to
retrieve similar source-language examples and use source-language
prompting to improve performance on a target language. Finally,
\citewithname{tu-etal-2022-prompt} show that prompt-tuning
multilingual LLMs can outperform fine-tuning in a cross-lingual
setting. However, this previous work does not evaluate any approach on
emotion analysis.

\section{Experimental Setting}
\begin{figure}[t]
  \centering
  \includegraphics[scale=0.8]{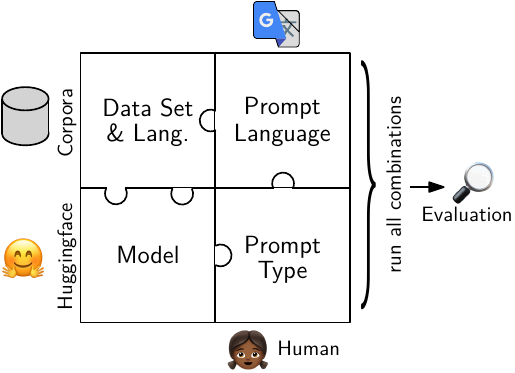}
  \caption{Overview of our experimental setting. We compare models
    from Huggingface and multiple prompt types for NLI-based emotion
    classification from
    \citewithname{plaza-del-arco-etal-2022-natural}. Across them, we study
    the relation between the data language and the prompt language for
    18 languages. To obtain the prompt in various languages, we apply
    Google Translate. An example setup would be the German subset of
    the Universal Joy corpus with an XLM-RoBERTa NLI model and the
    prompt as ``This person feels X'' translated to German (or left in
    English).}
  \label{fig:workflow-overview}
\end{figure}

For our experiments we use 6 multilingual NLI models, 3 emotion
corpora in 18 languages, and 7 prompt types. All experimentation is
performed in a zero-shot setting -- using no training or development
data. We explain the details in the following
section. Figure~\ref{fig:workflow-overview} depicts this setup.

\subsection{Data}
We use three different emotion corpora which
combine multiple languages.
The \textit{de/enISEAR} corpora \citep{Troiano2019} are manually
created emotion-triggering event descriptions collected by
crowdsourcing. The authors asked workers to describe an event that
caused in them a predefined emotion. It consists of 1001 instances for
both English and German, respectively, across 6 emotions.

\begin{table}[htbp]
  \centering
  \small
  \setlength{\tabcolsep}{3pt}
  \caption{List of languages used by Universal Joy (UJ) and more generally throughout the paper, sorted alphabetically by shorthand.}
  \begin{tabular}{ll|ll}
    \toprule
    Shorthand & Name & Shorthand & Name \\
    \midrule
    bn & Bengali & ms & Malay \\
    de & German & my & Burmese \\
    en & English & nl & Dutch \\
    es & Spanish & pt & Portuguese \\
    fr & French & ro & Romanian \\
    hi & Hindi & th & Thai \\
    id & Indonesian & tl & Tagalog \\
    it & Italian & vi & Vietnamese \\
    km & Khmer & zh & Chinese \\
    \bottomrule
  \end{tabular}
  \label{tab:languages}
\end{table}

The \textit{Universal Joy (UJ)} corpus
\citep{lamprinidis-etal-2021-universal} stems from Facebook posts in
18 languages (see Table~\ref{tab:languages} for a list).  The
motivation for creating this resource was to explore how emotions
manifest across languages. We use the predefined test split
(containing data for 5 comparatively higher resource languages),
downsampled to 981 instances for each of the languages. For the
remaining 13 languages (comparatively lower resource languages) there
is only one version of the dataset containing all their respective
instances. We subsample all of them to a maximum of 981 instances. The
data set contains 7 emotion categories.

The \textit{EmoEvent} corpus \citep{plaza-del-arco-etal-2020-emoevent}
consists of manually annotated Tweets in Spanish and English. We
remove all instances with the emotion labelled as `other' as well as
12 empty instances. This leads to 792 instances for English and 830
for Spanish across 7 emotions.

\subsection{Models}

We now describe the details of the 6 NLI models used for our
experiments, including which base language model was used and what NLI
dataset it was fine-tuned on.

\sloppy\paragraph{Natural Language Inference Datasets:} The NLI
datasets we use for fine-tuning are the Multi-Genre Natural Language
Inference corpus \citep[MNLI,][]{williams-etal-2018-broad}, the
Cross-lingual Natural Language Inference
corpus~\citep[XNLI,][]{conneau-etal-2018-xnli}, the Adversarial
Natural Language Inference
Dataset~\citep[ANLI,][]{nie-etal-2020-adversarial} and finally the
Tasksource dataset~\citep{sileo2023tasksource}. MNLI is a collection
of 433k English sentence pairs with entailment information, while XNLI
contains 7500 new English test examples following the annotation
procedure of \citewithname{williams-etal-2018-broad}, and then uses
manual translation to 15 languages in order to create a final dataset
of 112.5K combined development and testing examples.  ANLI is a
collection of NLI instances specifically designed to be difficult for
state-of-the-art models to solve, while Tasksource is a collection of
500 smaller datasets, including many for NLI.

\begin{table}
\caption{A list of the NLI models we use for our experiments. The names are links to the respective HuggingFace models. All of them have either a differing architecture or differing fine-tuning datasets to ensure a diverse sample of different models.}
\centering
\footnotesize
\begin{tabularx}{\linewidth}{llX}
\toprule
Name & Fine-tuned On & Base Model \\
\cmidrule(r){1-1}\cmidrule(rl){2-2}\cmidrule(l){3-3}
  \href{https://huggingface.co/vicgalle/xlm-roberta-large-xnli-anli}{\xlmroberta}
     & XNLI/ANLI & XLM-RoBERTa-large
  \\
  \href{https://huggingface.co/MoritzLaurer/multilingual-MiniLMv2-L6-mnli-xnli}{\minilm}
     & XNLI/MNLI & Distilled XLM-RoBERTa-large
  \\
  \href{https://huggingface.co/MoritzLaurer/ernie-m-large-mnli-xnli}{\ernie}
     & XNLI/MNLI & RoBERTa
  \\
  \href{https://huggingface.co/MoritzLaurer/xlm-v-base-mnli-xnli}{\xlmv}
     & XNLI/MNLI & XLM-V-base
  \\
  \href{https://huggingface.co/MoritzLaurer/mDeBERTa-v3-base-mnli-xnli}{\mdeberta}
     & XNLI/MNLI & mDeBERTa
  \\
  \href{https://huggingface.co/sileod/mdeberta-v3-base-tasksource-nli}{\mdebertatasksource}
     & Tasksource & mDeBERTa (v3)
  \\
\bottomrule
\end{tabularx}
\label{tab:models}
\end{table}

\paragraph{Model Architectures:} We use pretrained multilingual
language models that have been fine-tuned on the NLI corpora described
above. This allows us to study the effects of \textit{model} and
\textit{prompt language} separately. If we instead used monolingual
models, these two variable always have to coincide, making it harder
to trace where an effect comes from.  In order to maximize the
generality of our claims, we sample a variety of model architectures
for our experiments.

Concretely, we experiment with: 
\begin{compactenum}
    \item a XLM-RoBERTa-large model fine-tuned on MNLI \& ANLI,
    \item a distilled version of XLM-RoBERTa-large (\minilm, \citealt{wang-etal-2021-minilmv2} fine-tuned on MNLI and XNLI,
    \item \ernie \cite{zhang-etal-2019-ernie}  fine-tuned on MNLI and XNLI,
    \item \xlmv \cite{2023arXiv230110472L} fine-tuned on MNLI and XNLI,
    \item \mdeberta \cite{he2021debertav3,he2021deberta} fine-tuned on MNLI and XNLI,
    \item and \mdebertatasksource, which has been fine-tuned on the
      Tasksource dataset \cite{sileo2023tasksource}.
\end{compactenum}

We take the models from the Huggingface
Hub\footnote{\url{https://huggingface.co/models}} with all but
\xlmroberta and \mdebertatasksource being introduced by
\citet{laurer2022less}. The information on each \textit{model} can be
found in Table~\ref{tab:models}.

\subsection{Prompt Types}

To use NLI models in a zero-shot manner, we encode the data point we
want to classify as the premise and each of the possible labels (in
our case emotions) as the hypothesis and then choose the label with
the highest probability of being entailed by the premise.

To represent the labels, we use seven (of eight total\footnote{The
  original paper additionally uses a prompt that uses all synonyms for
  a particular emotion from the Emolex dictionary
  \cite{Mohammad2012b}. We omit this prompt due to computational
  constraints.}) \textit{prompt types} proposed by
\citewithname{plaza-del-arco-etal-2022-natural}. We define a prompt as
a mapping from the input text $x$ and emotion label $e$ to a template
$T$, where $T$ can be:

\begin{tabular}{ll}
  $T_\text{{Emo-Name}}$ & $x$: $e$ \\
  $T_\text{{Expr-Emo}}$ & $x$: This text expresses $e$ \\
  $T_\text{{Feels-Emo}}$ & $x$: This person feels $e$ \\
  $T_\text{{WN-Def}}$ & $x$: This person expresses $\text{wn}(e)$ \\
  $T_\text{{Emo-S}}$ & $x$: $syn(e)$ \\
  $T_\text{{Expr-S}}$ & $x$: This text expresses $\text{syn}(e)$ \\
  $T_\text{{Feels-S}}$ & $x$: This person feels $\text{syn}(e)$ \\
\end{tabular}

where $\text{wn}(e)$ is a function that maps an emotion to its WordNet
definition \cite{miller-1994-wordnet} and $\text{syn}(e)$ is a function that
maps an emotion to 6 predefined synonyms. For the all prompt templates
that use $\text{syn}(e)$, we run the model on all 6 prompts and take average
entailment probability as the final prediction.

We extend the prompts provided by
\citewithname{plaza-del-arco-etal-2022-natural} to cover the labels
\textit{anticipation} in UJ and \textit{surprise} in EmoEvent, and add
6 manually created synonyms for each.

\begin{table*}
  \centering \small
  \setlength{\tabcolsep}{3.6pt}
  \caption{Comparison (macro-\F across emotion categories) of the performance of using the English prompt
    for emotion classification or a translation to the data language
    (RQ1). The various scores are averages across
    \textit{prompt types} and NLI \textit{models}. EmoE: EmoEvent; de/enIS:
    en/deISEAR.}
  \begin{tabular}{lrrrrrrrrrrrrrrrrrrrrrrr}
    \toprule
    &\multicolumn{22}{c}{Dataset language} \\
    \cmidrule(){2-23}
    &\multicolumn{18}{c}{Universal Joy} &\multicolumn{2}{c}{EmoE} & \multicolumn{2}{c}{en/deIS} \\
    \cmidrule(r){2-19}\cmidrule(lr){20-21}\cmidrule(l){22-23}
    Prompt lang. & bn & de & en & es & fr & hi & id & it & km & ms & my& nl &pt& ro & th& tl &vi& zh &en& es &de &en\\
    \cmidrule(r){1-1}\cmidrule(lr){2-19}\cmidrule(l){20-21}\cmidrule(l){22-23}
    English & \textbf{25} & \textbf{26} & \textbf{30} & \textbf{26} & \textbf{28} & \textbf{21} & \textbf{26} & \textbf{24} & \textbf{25} & \textbf{26} & \textbf{25} & \textbf{27} & \textbf{31} & \textbf{27} & \textbf{27} & \textbf{26} & \textbf{28} & \textbf{34} & \textbf{32} & \textbf{13} & \textbf{23} & \textbf{29} \\
    Translated & 22 & 24 & --- & 24 & 26 & 19 & 24 & 23 & 23 & 25 & 19 & 24 & 28 & 26 & 25 & 20 & 27 & 31 & --- & 13 & 22 & --- \\
    \bottomrule
  \end{tabular}
  \label{tab:RQ1}
\end{table*}

We use Google Translate to obtain prompts in the 18 languages
of our data. 
Table~\ref{tab:examples} shows some
examples. We performed a manual analysis of the prompts in a subset of
the languages (German, Spanish) and confirm that the quality of
translation is generally high.

\begin{figure}[t]
    \centering
    \begin{subfigure}{\linewidth}
      \centering
      \caption{Universal Joy}
      \includegraphics[scale=\heatmapscale]{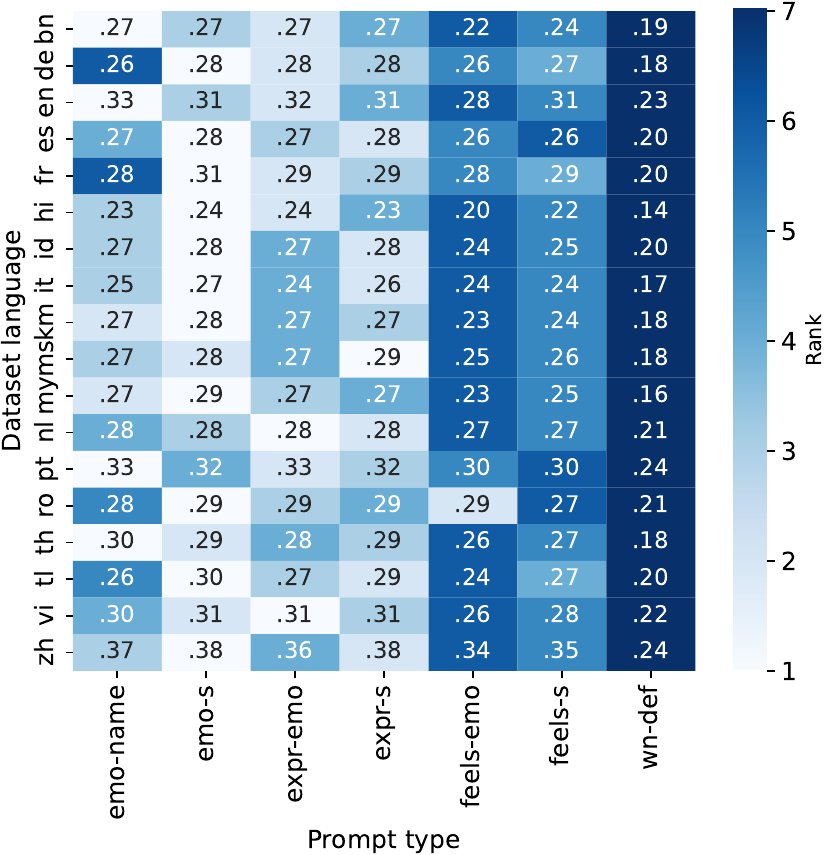}
      \label{fig:uj_heatmap}
    \end{subfigure}
    
    \begin{subfigure}{\linewidth}
      \centering
      \caption{de/enISEAR}
      \includegraphics[scale=\heatmapscale]{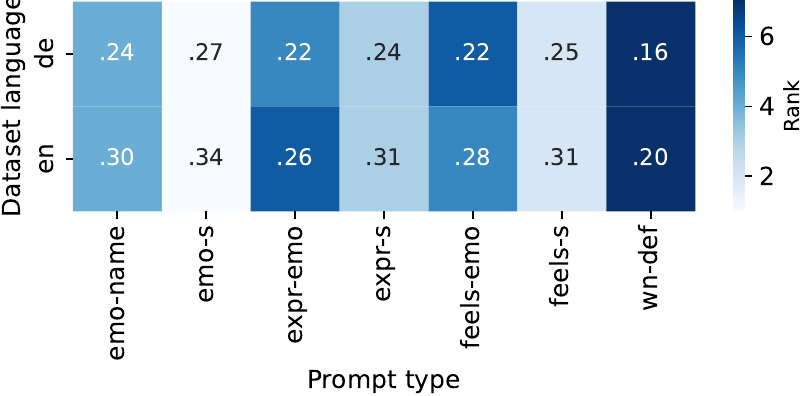}
      \label{fig:deisear_heatmap}
    \end{subfigure}

    \begin{subfigure}{\linewidth}
      \centering
      \caption{EmoEvent}
      \includegraphics[scale=\heatmapscale]{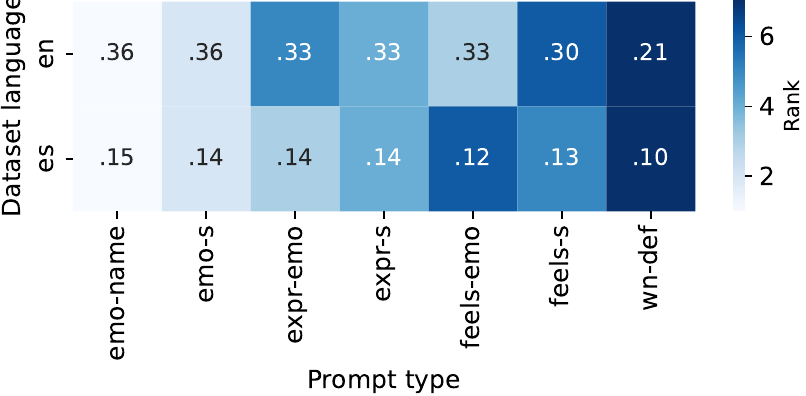}
      \label{fig:emoevent_heatmap}
    \end{subfigure}
    \caption{Interaction of \textit{prompt types} and \textit{data
        languages}. Each cell contains the average \F across NLI
      \textit{models}. The prompt is always in English. The color
      corresponds to the rank and therefore indicates consistency
of the results.}
    \label{fig:all_heatmaps_unmatched}
  \end{figure}

  \begin{figure}[t]
  \begin{minipage}{\columnwidth}
    \centering
    \subcaption{Universal Joy}
    \includegraphics[scale=\heatmapscale]{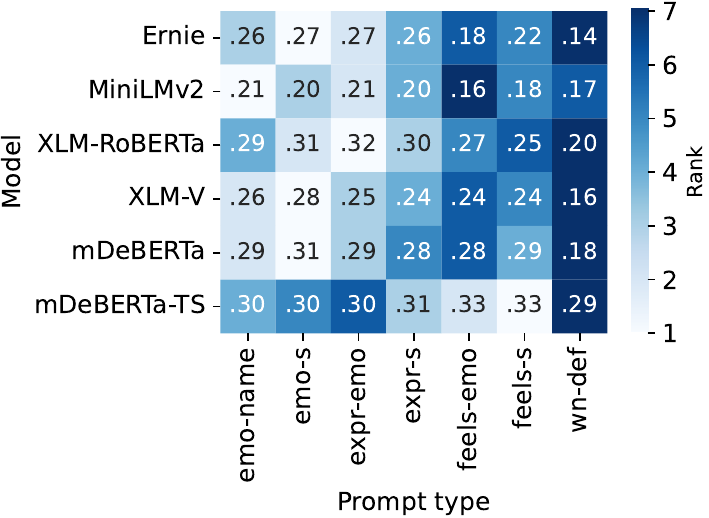}
    \label{fig:uj_model_consistency}
  \end{minipage}
  
  \begin{minipage}{\columnwidth}
    \centering
    \subcaption{de/enISEAR}
    \includegraphics[scale=\heatmapscale]{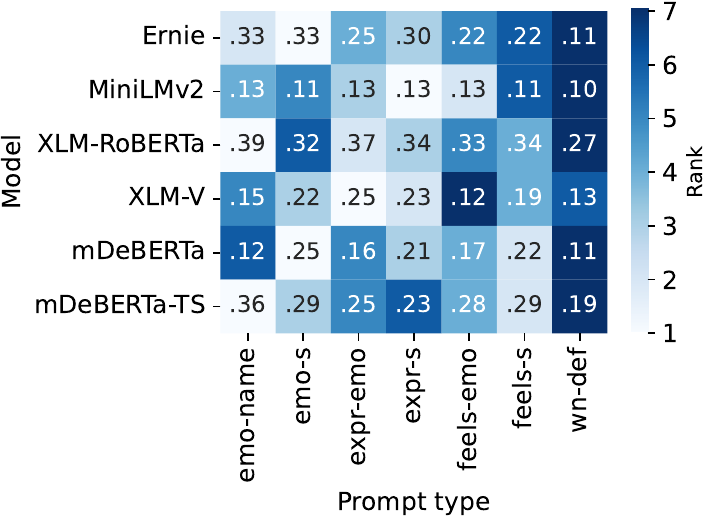}
    \label{fig:deisear_model_consistency}
  \end{minipage}

  \begin{minipage}{\columnwidth}
    \centering
    \subcaption{EmoEvent}
    \includegraphics[scale=\heatmapscale]{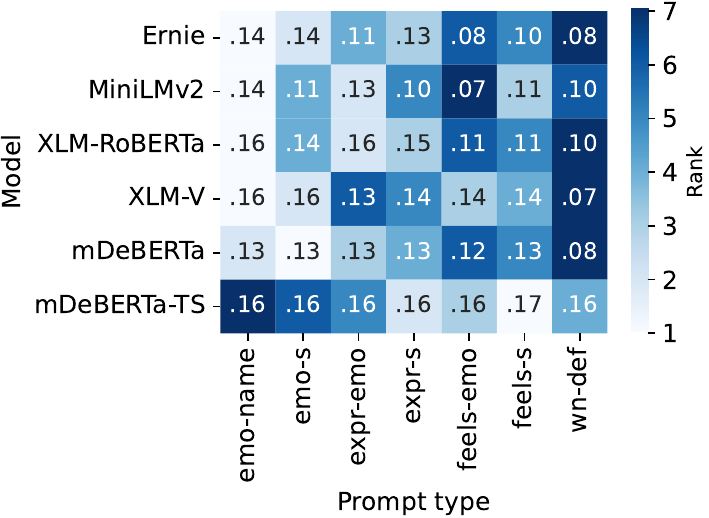}
    \label{fig:emoevent_model_consistency}
  \end{minipage}
    
  \caption{Interaction of \textit{NLI-models} and \textit{prompt
      types}. Cells are macro-average \F scores across \textit{prompt
      and data languages}. English data is omitted, as we are
    interested in the results on the target languages.}
    \label{fig:all_model_consistency_prompt_type}
\end{figure}

\subsection{Controlling for Variables of Interest }

Although it is in principle interesting to evaluate all possible
combinations of the four variables \textit{model}, \textit{data
  language}, \textit{prompt language} and \textit{prompt type}, due to
practical limitations, we restrict the \textit{prompt language} to
English and the translated target data language. This restriction is
motivated by the fact that English is well-represented in all training
sets of the \textit{models} we test. By matching the \textit{prompt
  language} and \textit{data language} via machine translation, on the
other hand, we capture a common use case in NLP. In total, we evaluate
1470 combinations for UJ and 126 for both de/enISEAR and EmoEvent this
way.

\section{Results}

Overall, models perform within the expected range for zero-shot
classification with a larger number of labels. Macro \F scores run
from 0.03--0.5, depending on the combination of model, prompt type, and
language. We therefore set out to answer the research question posed
in the introduction.

\subsection{RQ1: Should we translate the \textit{prompt language} to
  match the \textit{data language} or leave it in English?}

\paragraph{Overview.}
Multilingual NLI models can process prompts in either English or the
target language. It is reasonable to assume that the performance would
be higher if the data and prompt languages are the same. Here we test
this hypothesis.

\paragraph{Results.}
Table~\ref{tab:RQ1} shows the results of all models on the three
emotion corpora. The rows correspond to the prompt language (English
or translated to the data language) and the columns show the data
language. We report the macro \F scores for each emotion
classification setting, averaging over \textit{models}, \textit{prompt
  types}, and \textit{emotion label} for each target language in the
three data sets.

For some data sets and languages, the performance is lower than for
others, which we interpret as a varying difficulty of the respective
data sets. More interestingly for our RQ is that the English prompt
performance outperforms the target language prompts in all cases of
the Universal Joy Data Set (average \F difference of
0.025). For EmoEvent, the performance is roughly the same, while for
de/enISEAR, there is only a minor difference for the English--German
pair (of 0.013).

We therefore conclude that it is generally better or equally
beneficial to use an English prompt for performing emotion
classification in a target language. This observation is in line with
previous work
\citep{huang-etal-2022-zero,zhao-schutze-2021-discrete,etxaniz2023multilingual},
which finds that translating a prompt to a target language for other
tasks has no benefit and often directly harms model performance.

\subsection{RQ2: Is the performance of different \textit{prompt types} stable across different \textit{data languages}?}

\paragraph{Overview.}
Small variations to a prompt can lead to a drastic change in
classification performance
\citep{Liu2023,plaza-del-arco-etal-2022-natural}. Therefore, we ask
if if there is any concrete prompt type that performs particularly
well or poorly across all languages. Or instead, is the choice of
prompt type to use for emotion classification tied tightly to the
target language?

\paragraph{Results.}
We show the results in Figure~\ref{fig:all_heatmaps_unmatched} for the
three datasets. Each cell in Figure~\ref{fig:all_heatmaps_unmatched}
shows an average across \textit{models} for a combination of a
\textit{prompt type} (x-axis) and a \textit{data language}
(y-axis). The color in the heatmaps represents the rank of each prompt
type, \ie, the rank of the average performance for each prompt type
compared to the other 6 (for a given row, \ie, \textit{data
  language}). Given the results of RQ1 above, we fix the
\textit{prompt language} for this heatmap to be
English. 

Figure~\ref{fig:all_heatmaps_unmatched} indicates that the best
performing \textit{prompt types} are consistent across target
languages. The best-performing prompt for English data on UJ
(emo-name) is also in the top-3 \textit{prompt types} for 11 other
languages. The best overall \textit{prompt type} for other target
languages, however, is emo-s, which achieves the top ranking results
in 11 languages.  Wn-def is consistently the worst performing
\textit{prompt type}, followed by feels-emo and feels-s.
The results on EmoEvent and de/enISEAR are comparable to UJ.

To quantify the consistency across prompts in
Figure~\ref{fig:all_heatmaps_unmatched}, we calculate the average
Kendall's $\tau$ between all pairs of rows. The
correlation of different \textit{prompt types} between the languages
is .64 for UJ, .9 for de/enISEAR, and .62 for EmoEvent.

We conclude that there is a strong relation between the performance of
a prompt in English and the target languages. Therefore, we expect a
good prompt for English data to be good for data in other
languages. Similarly, we observe that prompt templates that ask the
model to estimate what a concrete actor is feeling (feels-emo,
feels-s) generally perform worse than others.

\begin{table}
    \centering
    \small
    \setlength{\tabcolsep}{2.5pt}
    \caption{Performance in macro-\F across emotion categories for the
      \textit{models} and \textit{prompt languages} in the Universal
      Joy data set. Each cell represents an average across
      \textit{prompt types} and \textit{data languages}. We average
      over the \textit{data languages}. English is omitted as we are
      mainly interested in consistency on low-resource languages.}
    \begin{tabular}{lcccccc}
        \toprule
        &\multicolumn{6}{c}{Prompt language} \\
        \cmidrule(l){2-7}
        & \multicolumn{2}{c}{UJ} & \multicolumn{2}{c}{de/enISEAR} & \multicolumn{2}{c}{EmoEvent} \\
        \cmidrule(r){2-3} \cmidrule(lr){4-5} \cmidrule(l){6-7}
        Model & en & transl. & en & transl. & en & transl. \\
        \cmidrule(r){1-1} \cmidrule(l){2-7}
        \ernie & \textbf{.25} & .21 & \textbf{.26} & .24 & .11 & \textbf{.11} \\
        \mdeberta & \textbf{.29} & .26 & .17 & \textbf{.18} & \textbf{.12} & .12 \\
        \mdebertatasksource & \textbf{.31} & .30 & .26 & \textbf{.28} & \textbf{.16} & .16 \\
        \minilm & \textbf{.21} & .17 & \textbf{.12} & .11 & \textbf{.11} & .10 \\
        \xlmroberta & \textbf{.28} & .27 & \textbf{.36} & .32 & \textbf{.14} & .12 \\
        \xlmv & \textbf{.25} & .23 & \textbf{.20} & .16 & \textbf{.13} & .13 \\
        \bottomrule
    \end{tabular}
    \label{tab:all_model_consistency_prompt_language}
\end{table}

\raggedbottom

\subsection{RQ3: How consistent are the results across different NLI \textit{models}?}

\paragraph{Overview.}
The NLI models we use vary in number of parameters, size, variety of
pretraining data, and NLI-datasets used for fine-tuning. Therefore, we
explore whether the effects found for RQ1 and RQ2 generally hold
across models. More specifically, we study whether the results for
prompt language or prompt type vary particularly for specific models.

\paragraph{Results -- Prompt type performance across models.}
Figure~\ref{fig:all_model_consistency_prompt_type} shows the relation
between \textit{model} and the \textit{prompt type}. Each cell in the
heatmaps shows an average across \textit{models} for a combination of a
\textit{prompt type} (x-axis) and a \textit{model} (y-axis). We are
interested in the performance consistency on low-resource languages
and therefore exclude English.  Similarly to the results above, the
rank shows the consistency of the performance of a prompt type across
models.

We see a high consistency across models, with the exception of
\mdebertatasksource. For most models, either emo-name or emo-s are the
best performing prompt types, while WN-def has the lowest or second
lowest performance across all models. The average correlations of the
performances for the prompt types across models is lower than across
for languages with 0.4 on UJ, 0.18 for de/enISEAR, and 0.23 for
EmoEvent.

This is mostly due to the outlier \mdebertatasksource. Omitting this
last row in the heatmaps from the correlation calculations leads to
0.7 on UJ, 0.52 for EmoEvent and 0.22 for de/enISEAR. We presume that
this is attributable to the use of the Tasksource dataset
\cite{sileo2023tasksource}, which is specific to this model.

Therefore, we conclude that the finding of RQ2 holds consistently on
the majority of models.

\paragraph{Results -- Prompt language performance across models.}
Finally, we show the results for both English and the translated
prompts across languages for all data sets in
Table~\ref{tab:all_model_consistency_prompt_language}. For all models,
leaving the prompt untranslated performs better on UJ and for the
majority of models on en/deISEAR and EmoEvent (4 out of 6 for both
cases), strengthening our results from RQ1.

Overall these results indicate that our findings on the superior
performance of English prompts from RQ1 are consistent across models.

\begin{table*}
    \centering
    \small
    \caption{Examples of predictions with English and German
      prompts. The model is \xlmroberta, the data is the German
      portion of de/enISEAR. The prompt is expr-emo. Correct
      predictions are printed in bold. The top part of the table shows
    examples where both the English and the German prompt lead to the
    same result, while the predictions differ in the bottom part.}
    \begin{tabu}{rlllll}
    \toprule
    & Sentence & True Label & English Prompt Pred. & German Prompt Pred. \\
    \cmidrule(rl){1-1}\cmidrule(rl){2-2}\cmidrule(rl){3-3}\cmidrule(l){4-4}\cmidrule(rl){5-5}
    1. & Ich fühlte ..., als ich Vater wurde. & Joy & \textbf{Joy} & \textbf{Joy} \\[-1.8mm]
    \rowfont{\tiny}
    & (I felt ... when I became a father.) & & (Prompt: This text expresses joy) & (Prompt: Dieser Text drückt Freude aus) \\
    2. & Ich fühlte ..., weil ich zu dick bin & Shame & Guilt & Guilt \\[-1.8mm]
    \rowfont{\tiny}
    & (I felt ... because I am too fat.) & & (Prompt: This text expresses guilt) & (Prompt: Dieser Text drückt Schuld aus) \\
    
    3. & Ich fühlte ..., als ein Onkel starb. & Fear & Sadness & Sadness \\[-1.8mm]
    \rowfont{\tiny}
    & (I felt ... when an uncle died.) & & (Prompt: This text expresses sadness) & (Prompt: Dieser Text drückt Trauer aus) \\
    
    4. & Ich fühlte ..., als ich absagen musste & Sadness & Shame & Shame \\[-1.8mm]
    \rowfont{\tiny}
    & (I felt ... when I had to cancel.) & & (Prompt: This text expresses shame) & (Prompt: Dieser Text drückt Scham aus) \\
    
    5. & Ich fühlte ..., als ich geschwitzt habe & Shame & \textbf{Shame} & \textbf{Shame} \\[-1.8mm]
    \rowfont{\tiny}
    & (I felt ... when I sweated.) & & (Prompt: This text expresses shame) & (Prompt: Dieser Text drückt Scham aus) \\
    \cmidrule(rl){1-1}\cmidrule(rl){2-2}\cmidrule(rl){3-3}\cmidrule(l){4-4}\cmidrule(rl){5-5}
    6. &  Ich fühlte ..., als mein Hund krank war. & Fear & \textbf{Fear} & Sadness \\[-1.8mm]
    \rowfont{\tiny}
    & (I felt ... when my dog was sick.) & & (Prompt: This text expresses fear) & (Prompt: Dieser Text drückt Trauer aus) \\
    
    7. & Ich fühlte ..., als ich befördert wurde. & Joy & Shame & \textbf{Joy} \\[-1.8mm]
    \rowfont{\tiny}
    & (I felt ... when I got promoted.) & & (Prompt: This text expresses shame) & (Prompt: Dieser Text drückt Freude aus) \\
    
    8. & Ich fühlte ..., als der Urlaub vorbei war. & Sadness & \textbf{Sadness} & Joy \\[-1.8mm]
    \rowfont{\tiny}
    & (I felt ... when the vacation was over.) & & (Prompt: This text expresses sadness) & (Prompt: Dieser Text drückt Freude aus) \\
    
    9. & Ich fühlte ..., als ich vor ihrem Grab stand & Sadness & Shame & \textbf{Sadness} \\[-1.8mm]
    \rowfont{\tiny}
    & (I felt ... standing in front of her grave.) & & (Prompt: This text expresses shame) & (Prompt: Dieser Text drückt Trauer aus) \\
    
    10. & Ich fühlte ..., als ich schwer erkältet war. & Sadness & Fear & \textbf{Sadness} \\[-1.8mm]
    \rowfont{\tiny}
    & (I felt ... when I was severely cold.) & & (Prompt: This text expresses fear) & (Prompt: Dieser Text drückt Trauer aus) \\
    
    \bottomrule
    \end{tabu}

    \label{tab:examples}
\end{table*}

\section{Analysis}
To provide an intuition of the results, we show prompts with
predictions in Table~\ref{tab:examples}. We acknowledge that these
results are too few to gain any particular generalizable observations,
but hope that they still provide a better idea about how our methods
work and the results were obtained.

The top part of the table shows instances in which the English and the
translated prompt leads to the same predictions. Most instances
contain event descriptions that are clearly connotated with an
emotion. Becoming father (Example 1) is predominantly related to joy
and both the English and the German model infer this emotion to be
most appropriate. Similarly clear is the assignment of shame for the
event of sweating (Example 2). In Examples 2, 3, and 4, one might
argue that both labels are correct and the predicted labels are
acceptable labels for the text.

The lower part of the table shows instances in which the labels
inferred by the English and the translated prompt differ. As often the
case for prompt-based predictions, it is difficult to infer any
patterns from these instances. In Example 6 (description of a sick
dog), both fear (English prompt) and shame (German prompt) are
reasonable assignments. In Example 8 (vacations being over), the
German prompt is more prone to spurious correlations to the
associations of vacations with joy than the English prompt. English 7
(being promoted) and Example 9 (standing in front of a person's grave)
are challenging to interpret -- the labels predicted by the English
prompt make no sense compared to the German, data language, prompts.

We observe that there are indeed cases in which the data language
prompts outperform the English prompts, but there are also cases in
which the English prompts are less sensitive to potential biases in
underlying data. While these observations are hard to generalize,
given the few instances, they motivate future research which we will
mention in the next section.

\section{Conclusion and Future Work}
With this paper, we studied if English prompts for emotion
classification work well across various data languages and if the
results are robust to changes of the underlying language model and
reformulations of the prompt. We found that generally English prompts
outperform the prompts in the respective data languages, and except
for one underlying model, they hold robustly across them.

Our main results support previous work that multilingual
language models often perform better on a task when the prompt is kept
in English, even for target languages that are typologically far from
English
\citep{huang-etal-2022-zero,zhao-schutze-2021-discrete,etxaniz2023multilingual}. This
suggests that multilingual models have an inherent bias towards
English, no matter what the target language is.

There are two exceptions to this general observations. First of all,
we only had one underlying language model that has been fine-tuned on
different NLI data. This model showed differing results for some
prompt types and therefore this variation on the setup requires more
future attention. It is important to better understand how the
training data of the language model and the prompt interact, and
particularly how this affects the transferability of prompts across
languages.

Secondly, we saw in the analysis that some instances do show more
reasonable results for target language prompts. While, overall, this
does not justify the use of target language prompts, understanding
better what such instances have in common might help to improve the
development of languages in other languages than English. This is
important for the majority of people who want to use multilingual
language models interactively but do not have a sufficient command of
English.

Further, we did focus on the setup in which the language model is
fixed and only the prompts receive variations. It may be assumed that
slightly adapting the language model to perform similarly on a target
language as it does on English could change the overall results and
enable other language prompts to perform comparably well. This
required approaches of cross-lingual model alignment under
consideration of specific prompts -- a research task that we are not
aware received any attention yet.

Additionally, in this paper, we concentrated on prompting for emotion
classification, where we predict a single label for each
text. However, emotion labels are not mutually exclusive. Therefore,
future work needs to also consider prompting for multilabel emotion
classification \citep{ohman-etal-2020-xed}. While a  
simple conversion of single labels to binary predictions would likely
lead to comparable results, models that can exploit label relations
might behave differently.

Finally and more broadly, future work could benefit from the
exploration of prompt-based cross-lingual transfer for less
restrictive styles of prompting as compared to ones based on NLI. For
instance, prompting-based on next-token prediction of autoregressive
languagel models like GPT-3 \cite{brown2020language} allows the
specification of (1) task instructions as well as (2) few-shot
examples, which is not easily possible for NLI-based prompting. The
impact of these features when choosing a prompt for cross-lingual
transfer is not well understood and will certainly benefit from
additional work.

\bibliographystyle{ACM-Reference-Format}
\bibliography{lit.bib}

\appendix

\end{document}